\documentclass[graybox]{sty/svmult}

\usepackage{mathptmx}       
\usepackage{helvet}         
\usepackage{courier}        
\usepackage{type1cm}        
\usepackage{booktabs,
            makecell,
            tabularx,
            threeparttable}
\usepackage{booktabs}
\usepackage{siunitx} 
\usepackage{changepage}
\usepackage{hyperref}
\usepackage{makeidx}         
\usepackage{graphicx, subcaption}        
\usepackage{multicol}        
\usepackage{multirow}
\usepackage[table,xcdraw]{xcolor}

\makeindex             

\begin{document}

\title*{Commonsense Generation and Evaluation for Dialogue Systems using Large Language Models}
\author{Marcos Estecha-Garitagoitia and Chen Zhang and Mario Rodríguez-Cantelar and Luis Fernando D'Haro}
\institute{Marcos Estecha-Garitagoitia \at ETSI de Telecomunicación, Universidad Politécnica de Madrid, Av. Complutense, 30, 28040, Madrid, Spain. \email{marcos.estecha.garitagoitia@upm.es}
\and Chen Zhang \at Human Language Technology Group at Electrical and Computer Engineering, National University of Singapore, Singapore, 119077. \email{chen_zhang@nus.edu.sg}
\and Mario Rodríguez-Cantelar \at Intelligent Control Group, Centre for Automation and Robotics UPM-CSIC, Universidad Politécnica de Madrid, C. José Gutiérrez Abascal, 2, 28006, Madrid, Spain. \email{mario.rcantelar@upm.es}
\and Luis Fernando D'Haro \at ETSI de Telecomunicación, Universidad Politécnica de Madrid, Av. Complutense, 30, 28040, Madrid, Spain. \email{luisfernando.dharo@upm.es}}

\authorrunning{Estecha-Garitagoitia et al.}
\titlerunning{Commonsense Synthesis using Large Language Models}
%
%
\maketitle

\abstract{This paper provides preliminary results on exploring the task of performing turn-level data augmentation for dialogue system based on different types of commonsense relationships, and the automatic evaluation of the generated synthetic turns. The proposed methodology takes advantage of the extended knowledge and zero-shot capabilities of pretrained Large Language Models (LLMs) to follow instructions, understand contextual information, and their commonsense reasoning capabilities. The approach draws inspiration from methodologies like Chain-of-Thought (CoT), applied more explicitly to the task of prompt-based generation for dialogue-based data augmentation conditioned on commonsense attributes, and the automatic evaluation of the generated dialogues.\newline\indent
To assess the effectiveness of the proposed approach, first we extracted 200 randomly selected partial dialogues, from 5 different well-known dialogue datasets, and generate alternative responses conditioned on different event commonsense attributes. This novel dataset allows us to measure the proficiency of LLMs in generating contextually relevant commonsense knowledge, particularly up to 12 different specific ATOMIC \cite{hwang-2021} database relations. Secondly, we propose an evaluation framework to automatically detect the quality of the generated dataset inspired by the ACCENT \cite{sarik-2023} metric, which offers a nuanced approach to assess event commonsense. However, our method does not follow ACCENT's complex event-relation tuple extraction process. Instead, we propose an instruction-based prompt for each commonsense attribute and use state-of-the-art LLMs to automatically detect the original attributes used when creating each augmented turn in the previous step.\newline\indent  
Preliminary results suggest that our approach effectively harnesses LLMs capabilities for commonsense reasoning and evaluation in dialogue systems.}

\keywords{commonsense knowledge, event commonsense reasoning, natural language generation, prompt-based generation and evaluation, open-domain dialogues}

\section{Introduction}
\label{sec:intro}

The challenge of instilling commonsense intelligence in AI systems, especially in Natural Language Processing (NLP), represents a significant yet challenging goal. Human-like reasoning about everyday situations and events remains a complex task for AI systems~\cite{richardson-2023}. In this context, Large Language Models (LLMs), such as ChatGPT~\cite{openai-2022} and GPT-4~\cite{openai-2023}, have emerged as pivotal enablers in this domain \cite{li-2022}. Their robust zero-shot and few-shot learning capabilities have demonstrated substantial improvements across various NLP tasks~\cite{kocon-2023}. Due to their general-purpose design, they are easy to adapt to downstream tasks, endowing them with the necessary background knowledge of the physical and social world. Beyond the conventional approach of model fine-tuning where all model parameters are adjusted~\cite{howard-ruder-2018}, in-context learning methods have been shown to achieve high performance in modulating LLM’s behavior through prompts~\cite{liu-yuan-2021}.



AI's commonsense reasoning has evolved from the structured logic ~\cite{speer-2017} ~\cite{hwang-2021} to more context-rich approaches ~\cite{ghosal-cider-2021} ~\cite{ghosal-cicero-2022}. Former approaches primarily structure commonsense knowledge in logic-based formalisms, while the latter approaches emphasize in natural language-based inferences for dialogue reasoning. Aproaches such as COMET~\cite{bosselut-comet-2019} focuses on generating novel commonsense knowledge by adapting pretrained neural language models to create new knowledge tuples. Likewise, our preliminary research pivots around the concept of redefining commonsense knowledge representation by developing a dataset using a prompt-based approach. Moving away from conventional knowledge graph and logical expression methodologies, we propose to harness the expansive potential of LLMs. Using the relational framework of the ATOMIC\cite{hwang-2021} dataset as a grounding, our goal is to generate new dialog turns in natural language. These turns are designed to be instilled with commonsense knowledge that is contextually relevant and nuanced, reflecting commonsense relational insights. This approach not only promises a more human-like interaction in AI-driven dialogues, but also aligns with the natural progression of AI towards a more intuitive understanding of human communication~\cite{choi-2022}.

Based on the above, the motivation is driven by the preliminary objective of exploring the feasibility of using these language systems to evaluate the common sense obtained from automatic generation systems.
The effectiveness of our approach is assessed using a framework inspired by ACCENT~\cite{sarik-2023}. Unlike ACCENT's approach to evaluate commonsense, our method concentrates on the direct generation and evaluation of commonsense content by LLMs. For that purpose, we harness from ATOMIC's relational framework.




The main contributions of this paper are:

\begin{itemize}
    \item Design and generate an event commonsense expanded dataset using LLMs specifically targeting open-domain dialogues.
    \item Propose an evaluation methodology to automatically assess the quality of the generated expanded dataset by considering each commonsense attribute.
    \item Provide detailed information on the prompt templates used to generate and evaluate the commonsense synthetic dataset.
\end{itemize}

\section{Background}
\label{sec:background}

In recent years, the goal of making machines manifest commonsense understanding and capabilities like humans has been an active field for conversational AI. Although early research looked at how machines understand basic facts and ideas~\cite{mccarthy-hayes-1981}~\cite{lecun-2021}~\cite{lenat-1995} and ~\cite{liu-singh-2004}, a new approach has emerged by means of LLMs. Building upon the available Commonsense Knowledge Bases (CSKB), LLMs have enabled to foster less abstract forms of representing social commonsense knowledge. ConceptNet~\cite{speer-2017} and ATOMIC~\cite{hwang-2021} have emerged as widely used resources for dataset curation, establishing a de facto standard.

Our exploratory research takes ATOMIC as foundation and examines two intersectional areas in conversational AI: the synthesis of commonsense datasets via prompt-based processing, and the automatic quality assessment of the generated commonsense content. This dual focus addresses the critical need for not only generating contextually rich commonsense knowledge, but also ensuring its alignment with the intended parameters of conversational AI systems. Our work aims to refine and advance these mutually linked aspects, thus contributing to the development of more nuanced and reliable conversational agents.


\runinhead{Commonsense Dataset Synthesis} 
Various works has already addressed the issue of curating commonsense reasoning from foundational commonsense symbolic datasets. Ghosal~\cite{ghosal-cider-2021} presented CIDER, a dataset marked by commonsense inferences in dialogues. This work was further developed with CICERO~\cite{ghosal-cicero-2022} in 2022. CICERO expanded the event-based reasoning annotations for better commonsense inference in dialogue systems. Different researchers have delved into particular kinds of commonsense such as exploring temporal aspects~\cite{qin-2021} or examining ethical dimensions~\cite{ziems-2022}~\cite{hyunwoo-2022}~\cite{hao-2022}.

Beyond the existing methodologies based on using and filtering of pre-existed datasets, SODA~\cite{hyunwoo-soda-2022} opened a new stream of research. Its approach aims at distilling new generalist dialogues from the implicit commonsense knowledge that is already embedded in LLMs. This synthesis of dialogues seeks to overcome the existing limitations in generation, evaluation and understanding due to the dependence on these small datasets. Along these lines, SYNDICOM~\cite{richardson-syndicom-2023} further pushes these ideas proposing a methodology based on the incorporation of valid and invalid answers with feedback to dialogue contexts. Its methodology aims to improve the predictive capacity of the system's dialogues. These responses are obtained from templates created with the logical formalisms contained in ATOMIC and chaining these inferences together to form multiple dialogue turns.

Different to previous research, we propose a synthesis methodology at turn-level. Relying on the use of several open-domain databases, we expand the implicit knowledge of the following turn in a contextualized dialogue by considering up to 12 different commonsense dimensions. The advantage is that we are able to provide new alternative turn responses that focuses on a particular commonsense aspect. 

\runinhead{Commonsense Evaluation with LLMs}
Thanks to the wide general-purpose capabilities that LLMs provide, their use have been spread to all research layers. The strategies followed to evaluate vary in both procedure and application. This paper focuses on exploring their performance as zero-shot classification systems, where no task-specific supervised training is provided to the model. Our proposal builds on a number of recent works~\cite{ronak-203}~\cite{sun-2023}~\cite{zhen-qin-2023}~\cite{ma-2023} that directly use LLMs as classifiers in a multi-stage classification process, where the main focus is on prompt-engineering to perform the classification task. Although there is much effort towards open-sourcing competitive LLMs, OpenAI’s ChatGPT and GPT-4 remain the most successful LLMs to date~\cite{wang-2023}, and therefore we used them in our experiments. 

Similar to our approach, SODA~\cite{hyunwoo-soda-2022} uses GPT-3.5 to automatically validate the type of commonsense found in conversations that had previously been generated from one of ATOMIC's commonsense attributes. ACCENT~\cite{sarik-2023} explicitly focuses on computing continuous commonsense alignment scores by relying on the use of an event extraction model and a generative commonsense model. Taking these works as a starting point, our idea follows these same lines of commonsense contradiction detection, but extending the turn-level commonsense dimensions to develop fine-grained detection. Finally, we also include the usage of ACCENT as a means to compare the results obtained with LLMs.

\section{Methods}
\label{sec:methodology}

Inspired by recent prompt-based commonsense synthesis methods, we describe a method to expand and evaluate commonsense at turn-level for dialogue systems.

\subsection{Commonsense Knowledge Graph}
\label{subsec:commonsense-knowledge-graph}

Our idea is to expand the turns of given dialogues along certain given commonsense knowledge dimensions. For this research, we selected ATOMIC as the reference CSKB. It includes knowledge about physical entities, events-centered and social interactions. Its components focused on events and social interactions correspond to 84.4\% over the entire CSKB and provide information on how human events/actions interact with other events/actions. We considered these relationships as a reference for the definition of our dimensions that we want to expand in each turn. Table \ref{tab:relation-definitions} shows our definitions for each type of symbolic relation taken from ATOMIC.

\subsection{Source and Test Data}
\label{subsec:source-datasets}

To build our experimental dataset we use dyadic dialogues from five datasets: DailyDialog~\cite{li-etal-2017-dailydialog}, Topical-Chat~\cite{gopalakrishnan2019topical}, EmpatheticDialogues~\cite{rashkin-etal-2019-towards}, Persona-Chat~\cite{zhang-etal-2018-personalizing, dinan2019second} and Wizard Of Wikipedia~\cite{dinan2019wizard}.

\textcolor{black}{
For each dataset, 40 random dialogues with 5 to 10 turns were selected. These dialogues were expanded using the GPT-3.5 model; for the GPT-4 model, only 20 of the 40 selected dialogues were used. The decision to choose a lower number of dialogs to be expanded with GPT-4 is primarily for economic reasons, given the preliminary nature of this paper. Moreover, it is also assumed that, in general, GPT-4 performs better than GPT-3. Under these assumptions, we decided to halve the number of dialogs expanded with GPT-4. After this selection process, 1,721 turns were augmented a total of 12 times each turn, once for each dimension in Table \ref{tab:relation-definitions}.
}

\textcolor{black}{
The augmented responses obtained by the LLM were on average 35\% longer than the original sentence, showing that the new responses incorporate more information than the original response (probably due to the incorporation of commonsense information, but also a typical characteristic of the LLMs in spite of being requested to be concise in the prompt instruction).
}


\subsection{Methods and Prompt Design}
\label{subsec: prompt-design}

Recent work has shown that zero-shot listwise prompt-based give better results for classifying several text fragments simultaneously~\cite{ma-2023}~\cite{zhen-qin-2023}. On this basis, we pose that, despite the fact that these statements have been claimed to classify natural language text, similar results can be obtained for data synthesis by following an equivalent approach. In our case, the synthesis of the responses at turn-level by inducing given definitions of each knowledge commonsense dimension.

Therefore, our problem is stated: given a set of commonsense attribute definitions $R = \{ r_1, ..., r_M\} $, a dialogue context $ D =\{t_{i-n}, ..., t_{i-1}, t_i\} $ and the task description $p$, the job of the language model is to return a list of responses $ X = \{x_1, ..., x_M\}$ at turn $t_{i+1}$ following each attribute definition. Each response $x_j$ aligns with the definition of the corresponding commonsense attribute $r_j$ for $j = 1, 2, ..., M$.

\begin{figure}[htp]
    \centering
    \includegraphics[width=1.1\textwidth]{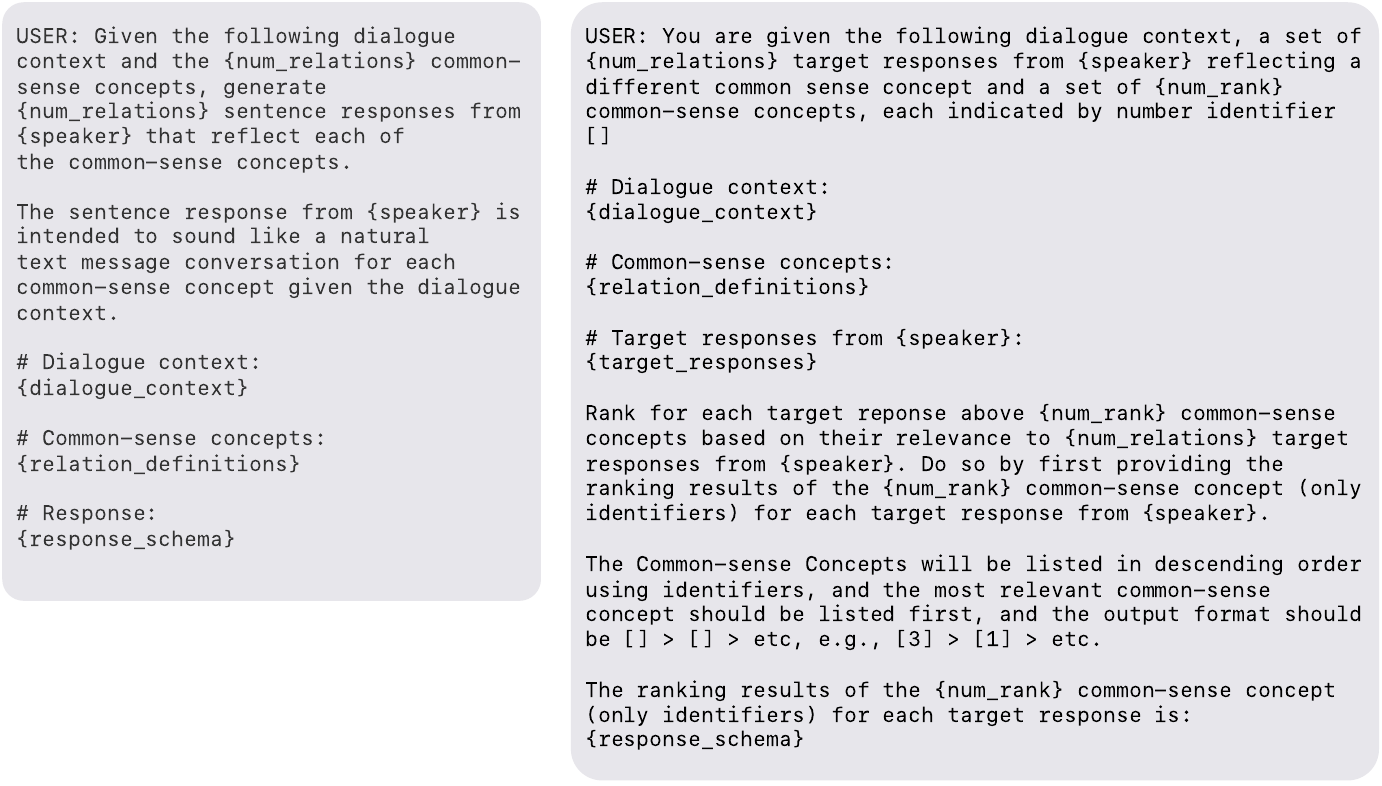}
    \caption{Expansion (Left) and Evaluation (Right) prompt templates.}
    \label{fig:expansion-instruction-template}
\end{figure}

For the prompt used in the evaluation, we have taken as a reference the prompts proposed in RankGPT~\cite{sun-2023} and adapted it to the commonsense task; specifically, the description stated in the prompt is different. The exact input prompts for generation and evaluation are shown in Figure \ref{fig:expansion-instruction-template}.



\begin{table}[ht]
    \caption{Human readable definition templates for each relation used for automatic commonsense expansion.}
    \label{tab:relation-definitions}
    \def\arraystretch{1.2}
    \begin{adjustwidth}{-3.25cm}{0cm}
    \begin{tabular}{llll}
        \toprule
        \textbf{Relations} & \textbf{Definitions} & \textbf{Relations} & \textbf{Definitions} \\ \midrule
        
        xAttr & \begin{tabular}[c]{@{}l@{}}The response should reflect what \{support\_speaker\} looks like \\ after going through what is being talked about. \{example\}\end{tabular} & oWant & \begin{tabular}[c]{@{}l@{}}The response should reflect the final objective \{speaker\} \\ desires to reach following the conversation. \{example\}\end{tabular} \\ \midrule
        
        xWant & \begin{tabular}[c]{@{}l@{}}The response should reflect the final objective \{support\_speaker\} \\ desires to reach following the conversation. \{example\}\end{tabular} & oReact & \begin{tabular}[c]{@{}l@{}}The response should reflect how \{speaker\} would react \\ to what is being talked about. \{example\}\end{tabular} \\ \midrule
        
        xNeed & \begin{tabular}[c]{@{}l@{}}The response should reflect the sequence of events or reasons \\ that need to happen prior to the conversation. \{example\}\end{tabular} & oEffect & \begin{tabular}[c]{@{}l@{}}The response should reflect how the situation will influences \\ \{speaker\} after the conversation. \{example\}\end{tabular} \\ \midrule
        
        xEffect & \begin{tabular}[c]{@{}l@{}}The response should reflect how the situation will influences \\ \{support\_speaker\} after the conversation. \{example\}\end{tabular} & HinderedBy & \begin{tabular}[c]{@{}l@{}}The response should state facts why what is being discussed \\ in the conversation could not happen. \{example\}\end{tabular} \\ \midrule
        
        xReact & \begin{tabular}[c]{@{}l@{}}The response should reflect how \{support\_speaker\} would \\ react to what is being talked about. \{example\}\end{tabular} & IsAfter & \begin{tabular}[c]{@{}l@{}}The response should reflect what led to the current situation \\ discussed with \{support\_speaker\}. \{example\}\end{tabular} \\ \midrule
        
        xIntent & \begin{tabular}[c]{@{}l@{}}The response should reflect what \{support\_speaker\} \\ wanted before the conversation. \{example\}\end{tabular} & HasSubEvent & \begin{tabular}[c]{@{}l@{}}The response should reflect the related causes and consequences \\ specific to the ongoing conversation. \{example\}\end{tabular} \\ \bottomrule
    \end{tabular}
    \end{adjustwidth}
\end{table}


The expansion process consists of generating new turns of dialogues following the commonsense given by the definitions of ATOMIC relations.

Although this paper explores zero-shot generation, we have also performed one-shot tests with examples generated automatically for each turn using COMET. To do so, we concatenate a specific example for each definition, replacing the {example} label shown in Table \ref{tab:relation-definitions}. The rationale behind this is to evaluate the performance when the system is induced with knowledge that could potentially be integrated versus self-generated knowledge. 

On the other hand, the evaluation process is defined as a listwise prompt-based classification process. Taking as ground-truths each relation that identifies the expansion of each turn, the GPT models are proposed to sort the definitions of the relations in the form of an ordered list. Thus, the task is to identify the definitions that best fit the expanded response. Once ranked, recommender system metrics are taken to evaluate the performance of the system for the evaluation task.



\section{Experiments}
\label{sec:evaluation}

To demonstrate the effectiveness of our proposal, we have evaluated the effectiveness of our approach by relying on the use of the ACCENT model and the GPT-3.5 and GPT-4 models to evaluate the previously expanded commonsense for each turn of the dialogues. GPT-3.5 refers to the model named gpt-3.5-turbo in the Open-AI suite, while GPT4 refers to gpt-4. 

Table \ref{tab:topk-mrr} compare the different configurations used to expand and evaluate the responses. Rows and columns identify expansion and evaluation processes followed, respectively. The ACCENT system has been implemented to evaluate in an integrated way. The other columns identify the prompted-based classification processes for each type of GPT model. As expected, all GPT-4 rerankers outperform the other benchmark methods.

\begin{table}[ht]
\caption{Top-k accuracy evaluation for each expansion method with k = \{1, 5, 10\} and Mean Reciprocal Rank (MRR) performance assesment.}
\setlength{\tabcolsep}{0.465em}
\begin{adjustwidth}{-3.5em}{0em}
\label{tab:topk-mrr}
\begin{tabular}{@{}lcrrrcrrrcccc@{}}
\toprule
 & \multicolumn{4}{c}{\textbf{ACCENT}} & \multicolumn{4}{c}{\textbf{GPT-3.5}} & \multicolumn{4}{c}{\textbf{GPT-4}} \\ \midrule
 & \multicolumn{3}{c}{\textbf{Top-k}} & \multicolumn{1}{c}{\multirow{2}{*}{\textbf{MRR}}} & \multicolumn{3}{c}{\textbf{Top-k}} & \multicolumn{1}{c}{\multirow{2}{*}{\textbf{MRR}}} & \multicolumn{3}{c}{\textbf{Top-k}} & \multirow{2}{*}{\textbf{MRR}} \\
 & \textbf{@1} & \multicolumn{1}{c}{\textbf{@5}} & \multicolumn{1}{c}{\textbf{@10}} & \multicolumn{1}{c}{} & \textbf{@1} & \multicolumn{1}{c}{\textbf{@5}} & \multicolumn{1}{c}{\textbf{@10}} & \multicolumn{1}{c}{} & \textbf{@1} & \textbf{@5} & \textbf{@10} &  \\ \midrule
\textbf{Zero-Shot GPT-3.5} & \multicolumn{1}{r}{0.08} & 0.37 & 0.64 & 0.245 & \multicolumn{1}{r}{0.20} & 0.54 & 0.85 & 0.367 & \multicolumn{1}{r}{0.21} & \multicolumn{1}{r}{0.58} & \multicolumn{1}{r}{0.88} & \multicolumn{1}{r}{0.388} \\ \midrule
\textbf{Zero-Shot GPT-4} & \multicolumn{1}{r}{0.08} & 0.38 & 0.65 & 0.249 & \multicolumn{1}{r}{0.19} & 0.52 & 0.86 & 0.355 & \multicolumn{1}{r}{\textbf{0.24}} & \multicolumn{1}{r}{\textbf{0.60}} & \multicolumn{1}{r}{\textbf{0.89}} & \multicolumn{1}{r}{\textbf{0.415}} \\ \midrule
\textbf{One-Shot GPT-3.5 w/Comet} & \multicolumn{1}{r}{0.09} & 0.39 & 0.65 & 0.255 & \multicolumn{1}{r}{0.19} & 0.53 & 0.86 & 0.362 & – & – & – & – \\ \bottomrule
\end{tabular}
\end{adjustwidth}
\end{table}

\begin{figure}[!ht]
  \begin{adjustwidth}{-12em}{}
  \begin{subfigure}{.8\textwidth}
    \centering
    \includegraphics[width=1\linewidth]{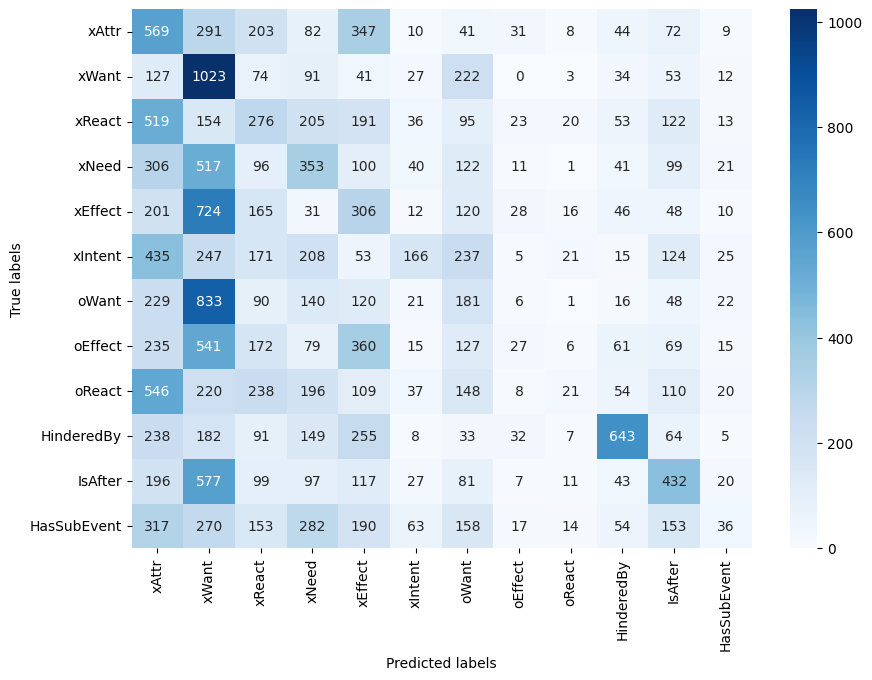}
    \caption{GPT-3.5 expansions ranked with GPT-3.5}
    \label{fig:gpt35-gpt35}
  \end{subfigure}%
  \begin{subfigure}{.8\textwidth}
    \centering
    \includegraphics[width=1\linewidth]{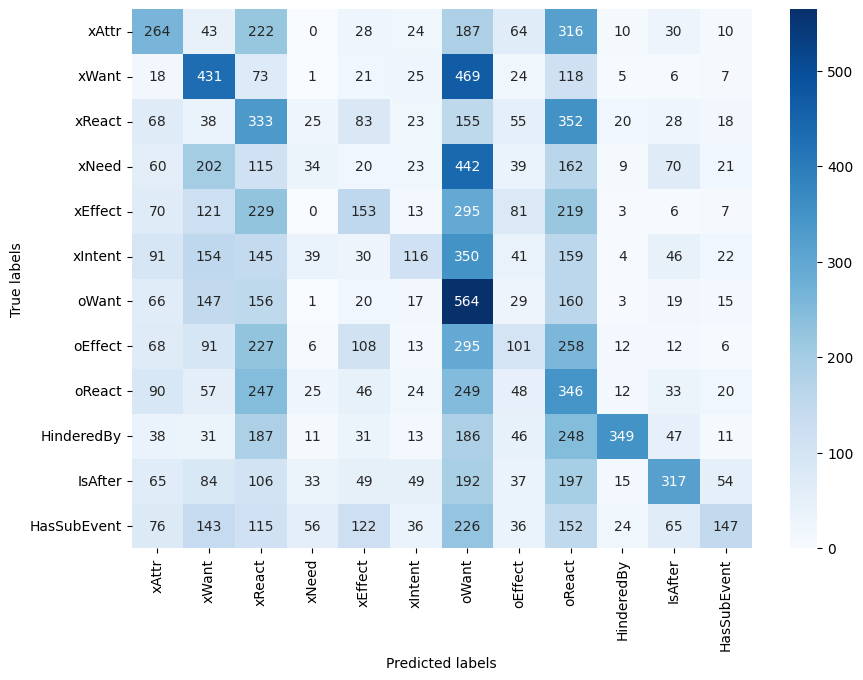}
    \caption{CGPT-3.5 expansions ranked with GPT-4}
    \label{fig:gpt35-gpt4}
  \end{subfigure}
  \medskip
  
  \begin{subfigure}{.8\textwidth}\quad
    \centering
    \includegraphics[width=1\linewidth]{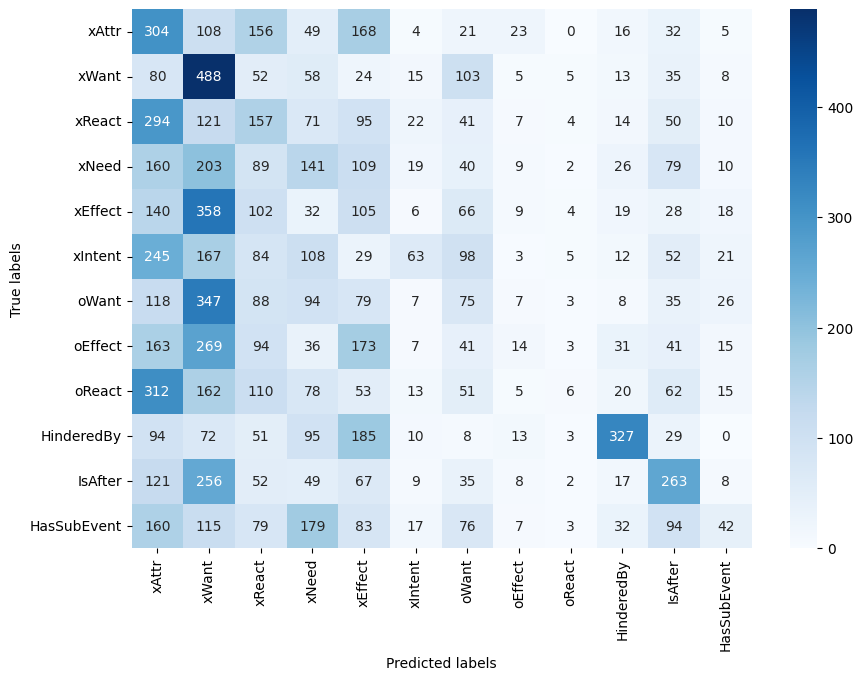}
    \caption{GPT-4 expansions ranked with GPT-3.5}
    \label{fig:gpt4-gpt35}
  \end{subfigure}%
  \begin{subfigure}{.8\textwidth}
    \centering
    \includegraphics[width=1\linewidth]{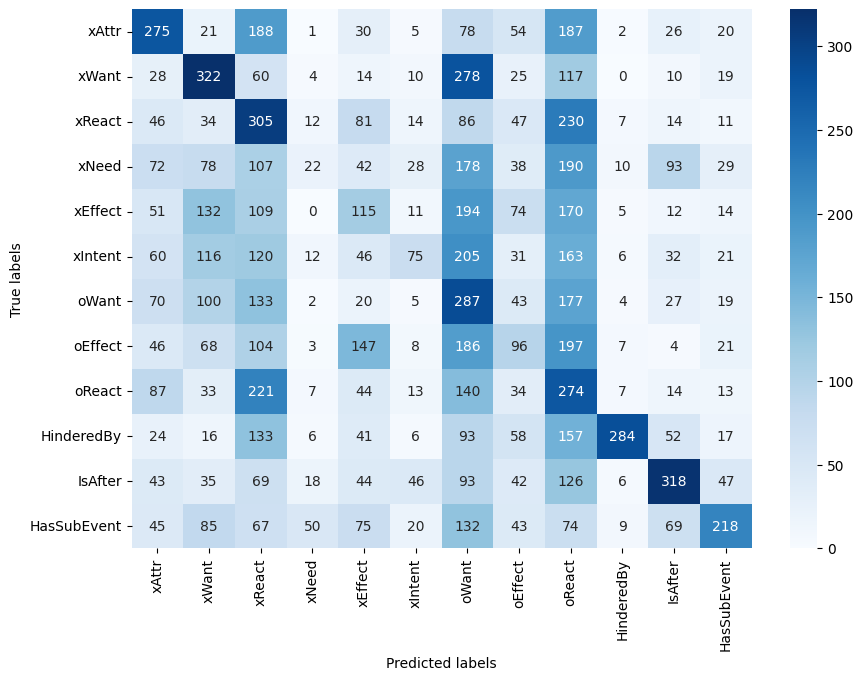}
    \caption{GPT-4 expansions ranked with GPT-4}
    \label{fig:gpt4-gpt4}
    
  \end{subfigure}
  \end{adjustwidth}
  \caption{Confusion matrices.}
  \label{fig:confusion-matrices}
\end{figure}

Figure \ref{fig:confusion-matrices} shows the results for the cases where we generated and cross-evaluated the GPT models. The diagonal is used to determine that GPT-4 is better suited to the task of identifying the target relation given the expanded responses. The confusion matrices for the case of GPT-3.5 are more sparse with certain biases towards certain relations. This may also be caused by an overlapping issue concerning the definitions of the commonsense relations. On the other hand, the cross-confusion matrices give us information about the quality of the sentences generated by GPT-3.5 and GPT-4. Taking into consideration Figures \ref{fig:gpt35-gpt4} and \ref{fig:gpt4-gpt4} we see how, despite having been generated by each of the models, the results are not very different and maintain a certain similarity in the values of the diagonal.


\section{Conclusions and Future Work}
\label{sec:conclusion}
In this study, we introduce a listwise zero-shot generation and reranking approach powered by open-source large language models. Our initial experiments show the viability of using LLMs for automatic generation and evaluation of commonsense dialogue turns. Both GPT-3.5 and GPT-4 generations were found very good (being GPT-3.5 a suitable option in case of considering costs), while for evaluation GPT-4 performed the best. 
In future work, we plan to increase the number of dialogues and perform manual annotations to better detect which commonsense attributes are better reflected in the expanded turns. Finally, we also plan to fine-tune a small generative model that can be used to automatically extract the attributes in chitchat dialogues.  

%
\begin{acknowledgement}
This work is funded by the European Commission through Project ASTOUND (101071191—HORIZON-EIC-2021-PATHFINDERCHALLENGES-01). In addition, this work is supported by project BEWORD (PID2021-126061OB-C43) funded by MCIN/AEI/10.1303-9/501100011033 and, as appropriate, by “ERDF A way of making Europe”, by the “European Union”; and by  Comunidad de Madrid: Research Grants for Young Investigators from Universidad Politécnica de Madrid (GENIUS:APOYO-JOVENES-21-TAXTYC-32-K61X37).
\end{acknowledgement}

\section*{Appendix}
\addcontentsline{toc}{section}{Appendix}

\begin{table}[]
\caption{Sample of expanded turn-level conversation along different commonsense dimensions.}
\label{tab:sample-1}
\begin{adjustwidth}{-7em}{}
\begin{tabular}{@{}clll@{}}
\toprule
\textbf{\#} & \textbf{Utterance} &  &  \\ \midrule
1 &
  \begin{tabular}[c]{@{}l@{}}User 1:  Well , what's the matter with you ?\\ GPT-4: I just had a really stressful \\ meeting earlier, and since then, I haven't \\ been able to shake off the negative feelings. \\ {[} cs: IsAfter {]}\end{tabular} &
  \begin{tabular}[c]{@{}l@{}}User 1:  Well , what's the matter with you ?\\ GPT-4: I've been dealing with some personal \\ issues lately, and it's been really getting me \\ down. I could use someone to talk to. \\ {[} cs: xAttr {]}\end{tabular} &
  \begin{tabular}[c]{@{}l@{}}User 1:  Well , what's the matter with you ?\\ GPT-4: I'm just feeling really down lately, \\ I think it's the stress catching up to me.\\ {[} cs: oReact {]}\end{tabular} \\ \midrule
2 &
  \begin{tabular}[c]{@{}l@{}}User 1:Well , what's the matter with you ?\\ User 2: I've got a sore throat and \\ my chest hurts .\\ GPT-4: Have you experienced any \\ other symptoms along with your sore \\ throat and chest pain, like fever or cough? \\ {[} cs: HasSubEvent {]}\end{tabular} &
  \begin{tabular}[c]{@{}l@{}}User 1: Well , what's the matter with you ?\\ User 2: I've got a sore throat and \\ my chest hurts .\\ GPT-4: Have you tried taking any \\ medicine or drinking warm tea with honey?\\ {[} cs: xNeed {]}\end{tabular} &
  \begin{tabular}[c]{@{}l@{}}User 1:Well , what's the matter with you ?\\ User 2: I've got a sore throat and \\ my chest hurts .\\ GPT-4: That doesn't sound good. You\\ should take it easy and not let it get worse.\\ {[} cs: oEffect {]}\end{tabular} \\ \midrule
3 &
  \begin{tabular}[c]{@{}l@{}}User 1: Well , what's the matter with you ?\\ User 2: I've got a sore throat and \\ my chest hurts.\\ User 1: Have you got a headache ?\\ GPT-4: Yeah, I have a headache too. \\ It's really affecting my ability to \\ focus on work and other activities.\\ {[} cs: HinderedBy {]}\end{tabular} &
  \begin{tabular}[c]{@{}l@{}}User 1: Well , what's the matter with you ?\\ User 2: I've got a sore throat and \\ my chest hurts.\\ User 1: Have you got a headache ?\\ GPT-4: I've been taking over-the-counter \\ meds, but they don't seem to be helping \\ much. Any suggestions?\\ {[} cs: xIntent {]}\end{tabular} &
  \begin{tabular}[c]{@{}l@{}}User 1: Well , what's the matter with you ?\\ User 2: I've got a sore throat and \\ my chest hurts.\\ User 1: Have you got a headache ?\\ GPT-4: No, I don't have a headache, \\ but I'm feeling quite weak and exhausted.\\ {[} cs: oWant {]}\end{tabular} \\ \midrule
4 &
  \begin{tabular}[c]{@{}l@{}}User 2: I've got a sore throat and \\ my chest hurts.\\ User 1: Have you got a headache ?\\ User 2: Yes , I ache all over .\\ GPT-4: That sounds like you might \\ have caught a virus or infection \\ recently. It'd be a good idea to see \\ a doctor to get it checked out.\\ {[} cs: IsAfter {]}\end{tabular} &
  \begin{tabular}[c]{@{}l@{}}User 2: I've got a sore throat and \\ my chest hurts.\\ User 1: Have you got a headache ?\\ User 2: Yes , I ache all over .\\ GPT-4: It's best to get plenty of rest \\ and drink fluids, but if your symptoms \\ persist or worsen, you should see a doctor.\\ {[} cs: xWant {]}\end{tabular} &
  \begin{tabular}[c]{@{}l@{}}User 2: I've got a sore throat and \\ my chest hurts.\\ User 1: Have you got a headache ?\\ User 2: Yes , I ache all over .\\ GPT-4: That sounds terrible. Are \\ you able to go about your day \\ with these symptoms?\\ \\ {[} cs: oEffect {]}\end{tabular} \\ \midrule
5 &
  \begin{tabular}[c]{@{}l@{}}User 1: Have you got a headache ?\\ User 2: Yes , I ache all over .\\ User 1: Are you coughing much ?\\ GPT-4: Besides the headache and cough, \\ I've also been having a fever and some \\ body aches. I guess I might have developed \\ a viral infection of some sort.\\ {[} cs: HasSubEvent {]}\end{tabular} &
  \begin{tabular}[c]{@{}l@{}}User 1: Have you got a headache ?\\ User 2: Yes , I ache all over .\\ User 1: Are you coughing much ?\\ GPT-4: Actually, the cough isn't too bad, \\ but it's making my headache worse.\\ {[} cs: oReact {]}\end{tabular} &
  \begin{tabular}[c]{@{}l@{}}User 1: Have you got a headache ?\\ User 2: Yes , I ache all over .\\ User 1: Are you coughing much ?\\ GPT-4:  Yeah, I've been coughing non-stop \\ and my headache is painful. I hope it's \\ just a cold or flu and not something \\ more severe.\\ {[} cs: HasSubEvent {]}\end{tabular} \\ \bottomrule
\end{tabular}
\end{adjustwidth}
\end{table}

\end{document}